\pdfoutput=1

\documentclass[11pt]{article}

\usepackage{acl}

\usepackage{times}
\usepackage{latexsym}
\usepackage{natbib}

\usepackage[T1]{fontenc}

\usepackage[utf8]{inputenc}

\usepackage{microtype}
\usepackage{graphicx}
\usepackage{chngpage}
\usepackage{multirow}
\usepackage{float}
\usepackage{nccmath}
\usepackage{amsmath}
\usepackage{amssymb}

\newcommand{\mymethod}{CERET\xspace}

\newcommand{\teal}[1]{\textcolor{teal}{#1}}

\title{CERET: Cost-Effective Extrinsic Refinement for Text Generation}

\author{
Jason Cai, Hang Su, Monica Sunkara, Igor Shalyminov, Saab Mansour \\ 
AWS AI Labs\\
 \small\texttt{\{cjinglun, shawnsu, sunkaral, shalymin, saabm\}}@amazon.com
 }

\begin{document}
\maketitle

\begin{abstract}
Large Language Models (LLMs) are powerful models for generation tasks, but they may not generate good quality outputs in their first attempt. Apart from model fine-tuning, existing approaches to improve prediction accuracy and quality typically involve LLM self-improvement / self-reflection that incorporate feedback from models themselves. Despite their effectiveness, these methods are hindered by their high computational cost and lack of scalability. In this work, we propose \mymethod, a method for refining text generations by considering semantic stability, entailment and inter-sample uncertainty measures. Experimental results show that \mymethod outperforms Self-consistency and Self-rerank baselines consistently under various task setups, by ~1.6\% in Rouge-1 for abstractive summarization and ~3.5\% in hit rate for question answering. Compared to LLM Self-rerank method, our approach only requires 9.4\% of its latency and is more cost-effective. \footnote{The source code and data samples are released at \url{https://github.com/amazon-science/CERET-LLM-refine}.}
\end{abstract}

\section{Introduction}

 Large Language Models (LLMs) like GPT \cite{brown2020language}, Claude, PaLM \cite{chowdhery2022palm, anil2023palm2}, and Llama \cite{touvron2023llama2} have showcased unprecedented capabilities in natural language understanding and generation. These models, with parameter counts reaching into the hundreds of billions, have become pivotal in advancing the frontier of natural language processing (NLP). Despite their impressive fluency and coherence, language models frequently generate content that is incomplete, biased, or misleading in their initial attempts across a variety of language generation tasks.

The key challenge is that while pre-training equips base models with broad linguistic knowledge, it does not necessarily impart the specialized skills needed for particular downstream tasks. 
Current methodologies for enhancing LLM generation largely involve resource-intensive approaches such as supervised fine-tuning (SFT), which relies heavily on domain-specific training data, or reinforcement learning from human feedback (RLHF), which necessitates extensive human annotations. However, curating large volumes of high-quality domain-specific data and human feedback often proves prohibitively expensive and time-consuming in practice, severely limiting the applicability of SFT and RLHF. 
By integrating feedback derived from the generated outputs, self-improvement / self-reflection approaches enhance generations in an iterative manner \cite{DBLP:journals/corr/abs-2303-17651,yao2023tree}. These approaches empower the LLM to adapt to specific tasks and domains by learning from its own mistakes and successes. Nevertheless, the substantial cost linked to iterative inference poses challenges for scalability and applicability real-time systems.

This paper introduces \mymethod, a novel method designed to refine text generation in a rapid, low-resource manner to reduce the need for domain-specific training data or expensive human annotations. The cornerstone of \mymethod lies in its ability to enhance generated content by holistically considering three key scoring dimensions - semantic stability, entailment, and inter-sample uncertainty measures.

Semantic stability scoring quantifies the linguistic invariance among multiple candidate outputs generated by the base model for the same input, indicating higher confidence for more stable candidates. Entailment scoring leverages natural language inference (NLI) models to quantify the logical entailment relations between candidate outputs, preferring candidates that maximally entail others. Inter-sample uncertainty scoring penalizes candidates that are semantically similar to outputs for different inputs, a signal of greater uncertainty.

Our approach operates in a rapid, zero-shot manner without any domain-specific training data, reward modeling, or human feedback. The proposed scoring and refinement process encapsulates an efficient way to improve text generation across a diverse spectrum of NLP tasks, including abstractive summarization, dialogue response generation, and open-domain question answering. Through a rigorous series of experiments on standard datasets, \mymethod  is empirically validated to significantly outperform baseline methods such as Self-consistency and Self-reranking across both summarization and QA tasks. Beyond its superior performance, \mymethod stands out for its practicality and cost-effectiveness, making it a promising solution for real-world applications where domain-specific resources and annotations are limited or unavailable. This paper not only presents \mymethod as a valuable novel contribution to the growing field of NLP but also underscores its potential impact on advancing the practical deployment of text generation across a myriad of domains.

The main contributions are summarized as follows:

\begin{itemize}
\item  \mymethod is proposed as a holistic framework for enhancing generation quality, encompassing semantic stability, entailment, and inter-sample uncertainty measures.

\item  The refinement process is data efficient and cost-effective, without the requirement for domain-specific training data or expensive annotations.

\item  The proposed approach can be applied across various natural language processing tasks, such as text summarization, dialogue response generation and question-answering systems.

\item  \mymethod is highlighted for its practicality and efficiency, presenting only a minor fraction of the usual latency associated with a single generation call, which positions it as a feasible solution for real-world applications.
\end{itemize}

\section{Approach}

\begin{figure*}[ht]
\centering
\includegraphics[width=16cm, height=8.5cm]{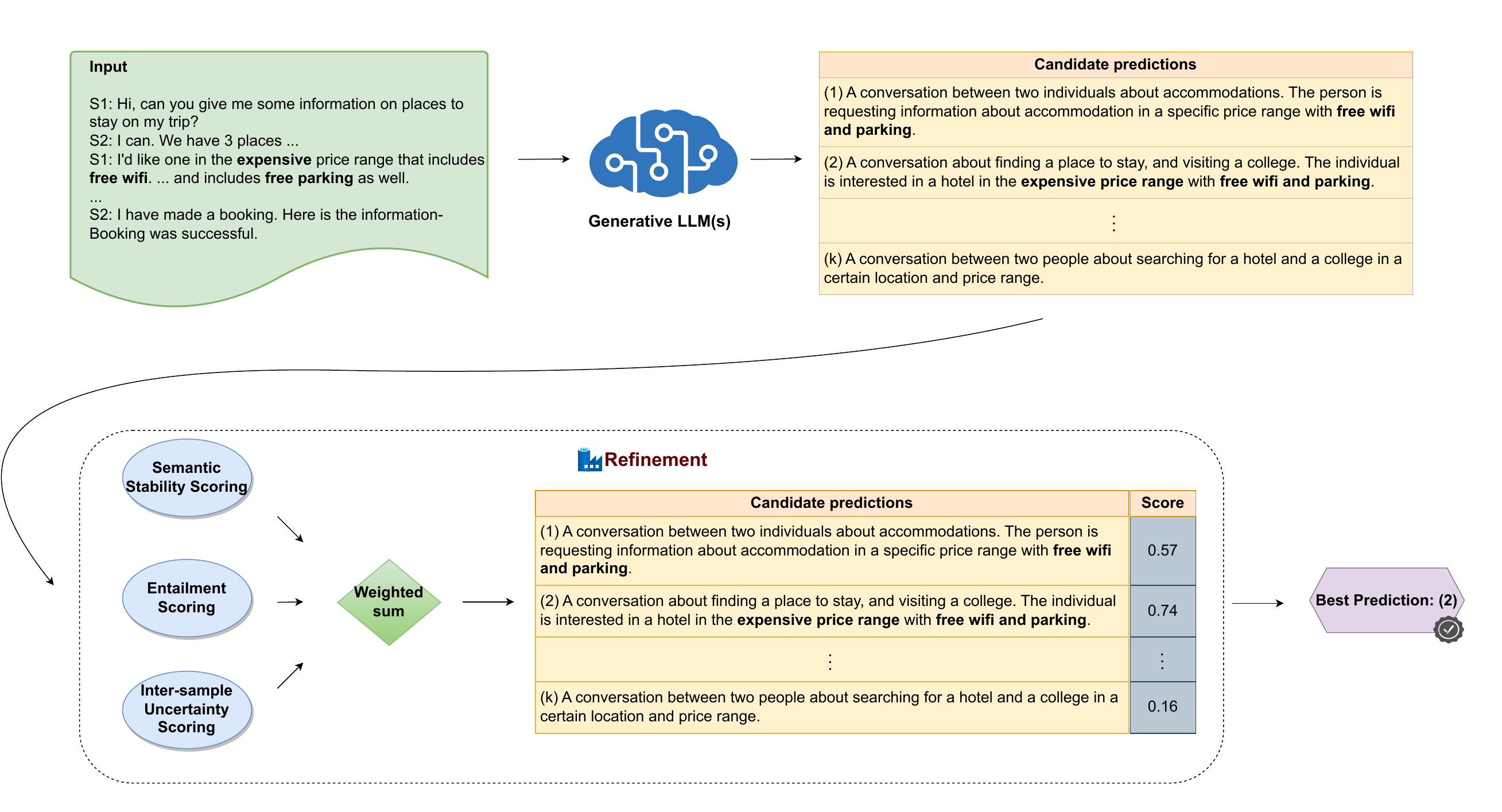}
\caption{\mymethod overview}

\label{fig_overview}
\end{figure*}

\subsection{System Architecture}

\mymethod consists of three scoring methods, namely Semantic Stability Scoring, Entailment Scoring and Inter-sample Uncertainty Scoring, for calibrating the quality of LLM predictions. The overview of the proposed system is illustrated in Figure~\ref{fig_overview}. Firstly, a diverse set of candidates are sampled from LLMs. Then each individual scoring method will produce a separate score from a certain perspective. Based on the scores in three dimensions, a linear weighted final confidence score is computed to measure the quality of each prediction. The prediction with the highest confidence score is selected as the final model prediction.

\subsection{Semantic Stability Scoring}

We first introduce an intra-sample scoring method, Semantic Stability Scoring, which is motivated by the need to enhance the confidence and reliability of sample generations produced by LLMs. The scientific rationale is inspired by \citet{DBLP:conf/iclr/KuhnGF23} and \citet{yin-etal-2022-sensitivity}, where it was shown that a sample generation exhibits higher confidence when it demonstrates considerable semantic stability or linguistic invariance among other generations. However, the semantic stability measured in \citet{DBLP:conf/iclr/KuhnGF23} involves clustering sampled generations for each sample, which is computationally expensive for real world applications at large scale. 

In contrast, we propose a cluster-free method for semantic stability modeling. Specifically, Semantic Stability Scoring is formulated as the following: Given input data sample \(x\), the model generates \(k\) predictions (\(y_1, ..., y_k\)). For each \(y_i\), a fixed pre-trained language model produces its corresponding embedding \(e(y_i)\). In practice, we leverage RoBERTa (A Robustly Optimized BERT Pretraining Approach) \cite{liu2019roberta} as the pre-trained language model, and the final hidden representation of ``<s>'' token from RoBERTa, is regarded as \(e(y_i)\). To aggregate all intra-sample representations, we treat the average-pooled embedding \(\bar{e}\) as a stability reference point:

\begin{equation}
\bar{e} = mean(e(y_1), ..., e(y_k))
\end{equation}

A lower distance between an embedding and the reference point implies a higher stability. We can employ Euclidean distance or cosine distance as the distance metric $||\cdot ||$. The stability score \(s_{sta}^{i}\) is defined as the negative distance between \(e(y_i)\) and the stability reference point \(\bar{e}\):

\begin{equation}
s_{sta}^{i} = - || e(y_i) - \bar{e} ||
\end{equation}

\subsection{Entailment Scoring}

Entailment scoring is another intra-sample scoring method, fully powered by entailment relation: ``p entails h'' (\(p \Rightarrow h\)), if a human reading premise \(p\) would infer that hypothesis \(h\) is most likely true. This intrinsic connection to human inference aligns closely with the objective that language models should have the capacity to generate content that is not only syntactically accurate but also semantically meaningful. In the entailment scoring process, when an LLM generates $k$ predictions (\(y_1, ..., y_k\)), each prediction's entailment relation to others is quantified by a Natural Language Inference (NLI) model. The scalar value $s^i_{j} $ reflects the degree to which the content of $y_i$ logically entails $y_j$.

\begin{equation}
s^i_{j} = \text{ENT}(y_i, y_j)
\end{equation}

Although the scalar function for entailment can be evaluated by the base LLM itself, such an approach leads to a higher computational cost. Hence, we resort to a more efficient and lightweight NLI model. Specifically, we adopt DeBERTa (Decoding-enhanced BERT with disentangled attention) \cite{he2021deberta} for this work. The NLI task is treated as a sequence classification problem: The texts $y_i, y_j$ are concatenated, with special tokens as separators, to form the input to DeBERTa. The final hidden representations of pretrained DeBERTa are passed to a pooling layer and a classifier, to obtain softmax probability for three categories, namely Neutral, Entailment and Contradiction. The softmax probability for Entailment is used as $\text{ENT}(y_i, y_j)$.

A generation is plausible if it entails as many other sampled generations as possible. With the top $k$ sampled model predictions, the entailment score for sample \(y_i\) is computed as follows:

\begin{equation}
s_{ent}^{i} = \frac{1}{k} \sum_{1 \leq j \neq i \leq k} s^i_{j} + LP(y_i)
\end{equation}

Note that the preferred prediction will likely have rich information, and may be lengthy in certain situations. A length penalty $LP(y_i)$ is applied to this entailment score, in case lengthy outputs harm the expected conciseness. 

\begin{equation}
LP(y_i) = 1 - (1 + q \cdot \text{len}(y_i)) ^ p
\end{equation}

where $0 \leq q <1$ and $p>1$ are hyperparameters\footnote{In our practice, we chose q = 0 (i.e. no penalty). We found that our beam search sampled predictions generally have very comparable lengths. Nevertheless, the length penalty may benefit other datasets or decoding settings.}.

\begin{figure}[ht]
\centering
\includegraphics[width=6.5cm, height=6.5cm]{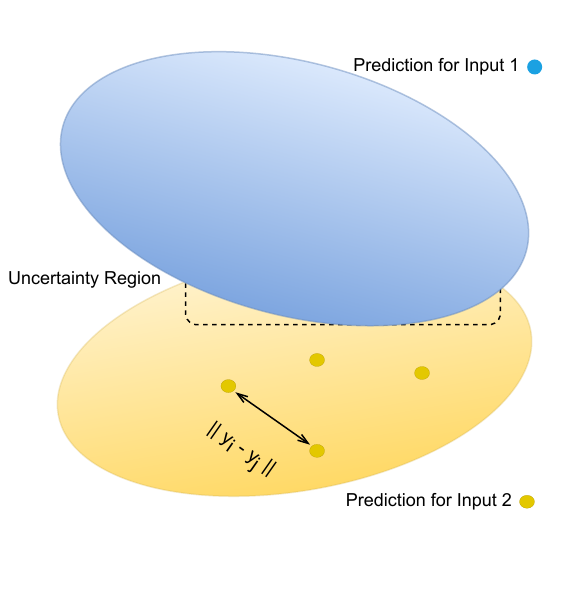}
\vspace{-5mm} 
\caption{Inter-sample uncertainty region}
\label{fig_unc}
\end{figure}

\subsection{Inter-sample Uncertainty Scoring}

In contrast to the methods above, the following is an inter-sample scoring method, which is inspired by uncertain region analysis. We first build an embedding space for all sampled predictions with a standalone model, e.g., RoBERTa. The rationale behind this inter-sample scoring method is that when a sampled prediction is located near predictions from different input data samples in the embedding space, this prediction is likely to be uncertain, illustrated in Figure~\ref{fig_unc}. Uncertain predictions are down-weighted by a lower uncertainty score \(s_{unc}\).

Suppose dataset \(D\) has size \(N\). For each input \(x\), top \(k\) predictions are generated by LLM(s), resulting in \(k \cdot N\) predictions in total: \(\{y_i\}_{1 \leq i \leq kN}\). As in Semantic Stability Scoring, we obtain the embeddings $e(y_i)$ from a pre-trained language model. The Euclidean distance of all possible prediction pairs \(\left\lVert e(y_i) - e(y_j) \right\rVert, i \neq j\) are computed and cached. According to Euclidean distance, the nearest neighbor set \(\mathcal{N}(i)\) is constructed for each prediction \(y_i\). The inter-sample uncertainty score \(s_{unc}^{i}\) is computed as follows:

\begin{equation}
s_{unc}^{i} = -\sum_{j \in \mathcal{N}(i)} \frac{\mathbb{I}(\hat{x}_i \neq \hat{x}_j)}{\left(1 + \left\lVert e(y_i) - e(y_j) \right\rVert\right)} 
\label{eq_unc}
\end{equation}

where \(\mathbb{I}(\cdot)\) is the indicator function. $\hat{x}_i$ denotes the input sample $x$ for prediction $y_i$. Note that possibly $\hat{x}_i = \hat{x}_j$ when $i \neq j$. $||e(y_i) - e(y_j)||$ in denominator of Equation~\ref{eq_unc} is a regularization term, ensuring a further $y_j$ is assigned with a lower weight for uncertainty. A negative sign is added to ensure that a higher score is better. In case when the dataset is large, the computation cost for obtaining pairwise Euclidean distances and nearest neighbors can be mitigated by limiting data size \(N\) to a certain batch (e.g., 1000). Additionally, the LLM generations in practice mostly have the number of sampled predictions \(k \leq 20\). Thus the efficiency of this method can be maintained.

\subsection{Computation of Final Score}

All separate scores \(s_{sta}\), \(s_{ent}\), \(s_{unc}\) are transformed to the interval \((0, 1)\) by applying sigmoid function 

\begin{equation}
\text{sigmoid} (x) = \frac{1}{1 + e^{-ux}}
\end{equation}
where \(u>0 \) is an additional scaling factor \footnote{For each scoring dimension, there is a dedicated value of $u$.}. Since three scores have distinct ranges, $u$ is applied to ensure their scaled ranges are comparable. The final confidence score is a linear weighted score based on three dimensions. 

\begin{equation}
s = \alpha \cdot s_{sta} + \beta \cdot s_{ent} + \gamma \cdot s_{unc}
\label{eq_weighted}
\end{equation}

The coefficients $\alpha, \beta, \gamma$ are tuned on validation datasets. To mimic the properties of probability for intuitive interpretation, the following constraints are imposed:

\begin{equation}
\left\{
\begin{aligned}
\alpha + \beta + \gamma & = 1 \\
\alpha, \beta, \gamma & \geq 0
\end{aligned} \right.
\end{equation}

\section{Experimental Setup}

\subsection{Datasets}
We evaluate the proposed approach \mymethod on Abstractive Summarization and Question Answering (QA) tasks. For summarization, we consider two dialogue summarization datasets: TodSum \cite{DBLP:journals/corr/abs-2110-12680} and DialogSum \cite{chen-etal-2021-dialogsum}, as dialogue summarization has been a challenging summaization use case due to its multi-speaker nature and varying structures.
TodSum is a dialogue summarization dataset based on MultiWoz \cite{DBLP:conf/emnlp/BudzianowskiWTC18}. Out of 7 MultiWoz domains, it contains 5 and totals 9,906 dialogues. DialogSum is a multi-domain dataset, mostly consisting of casual/spoken style daily conversations. It is based on top of the existing datasets around English practice conversations and English listening comprehension exams. For TodSum and DialogSum, official validation/test sets are used throughout this work.

For QA, we use TriviaQA \cite{DBLP:conf/acl/JoshiCWZ17} and Natural Questions \cite[the NQ-Open version]{lee-etal-2019-latent} datasets. TriviaQA contains 95,956 QA pairs with 40,478 unique answers and 662,659 evidence documents. It contains question-answer pairs from 14 trivia and quiz-league websites, with the associated Wikipedia pages as evidence sources. NQ-Open is an open-domain question answering benchmark, a subset of Natural Questions \cite{DBLP:journals/tacl/KwiatkowskiPRCP19} with short answers and with evidence documents discarded. It contains 91,535 QA pairs. For TriviaQA and Natural Questions, the official test set is only available for online benchmarking. We split the official validation sets into validation/test sets with an 1:1 ratio for our experiments.

\subsection{Baselines and Evaluation Metrics}
We choose Vicuna v1.3 \cite{vicuna2023} and Llama 2 chat \cite{touvron2023llama2} as our base LLMs. Vicuna is an open-source chatbot, fine-tuned from Llama with supervised instruction fine-tuning using around 125K conversations collected from ShareGPT. Llama 2 was pretrained on publicly available online data sources and trained on 2 trillion tokens,
and was initially created through supervised fine-tuning and then iteratively refined using Reinforcement Learning from Human Feedback (RLHF). Both Vicuna v1.3 and Llama 2 were released in mid 2023. 


Given each input prompt, we generate $k$ LLM predictions by beam search sampling \cite{vijayakumar2016diverse}, while setting a high temperature to encourage diversity and increase the scope for improvement. The beam search sampled predictions are considered as \textbf{No-refinement} baseline. We further considered two baselines. (1) \textbf{Self-rerank:} In the Self-rerank approach, all predictions generated by the base LLM and the task context are fed back into the base LLM itself. The model is then instructed to select the single best prediction from the candidate set. The Self-rerank baseline provides insight into the capabilities of the base LLM to refine its own output as a straightforward reranking task. (Prompt templates in Appendix~\ref{appendix_prompts}) (2) \textbf{Self-consistency:} The Self-consistency \cite{DBLP:conf/iclr/0002WSLCNCZ23} approach determines the best prediction through a majority vote among all generated predictions, after marginalizing out reasoning paths. This is a cost-effective approach for refinement, but it can only be applied to tasks with fixed answers. Hence, it is included as a baseline for open domain QA tasks. Furthermore, we also report \textbf{Oracle} scores, which represent the \textit{upper bound} of refinement/re-ranking performance: Given an input $x$, we obtain a candidate prediction set $\{y_i\}_i$, out of these $y_i$’s, we choose the best one according to certain evaluation metric (E.g., Rouge, Exact Match, etc) to compute oracle performance .

On the QA tasks with a closed set of answers, we evaluate the models against \textbf{Hit Rate}: A prediction receives score 1 if it exactly matches one of multiple target answers, otherwise score 0 is assigned. On the summarization tasks assuming more open-ended model outputs, we evaluate the models against \textbf{Rouge-1/2/L} \cite{lin-2004-rouge} and \textbf{BERTScore} \cite{DBLP:conf/iclr/ZhangKWWA20}. Rouge\footnote{\url{https://github.com/pltrdy/rouge}} is a series of metrics counting the number of overlapping word n-grams in the reference and the generated summary, working on top of 1-/2-grams (as the index in the metric name denotes). Rouge-L is a variant of the metric based on the Longest Common Subsequence between the reference and the generated summary. BERTScore\footnote{\url{https://github.com/Tiiiger/bert_score}} is a semantic similarity metric working in the BERT \cite{DBLP:conf/naacl/DevlinCLT19} embedding space by computing pairwise cosine similarities between each predicted summary's token and each reference summary's token.

\subsection{Implementation Details}
For the purpose of experimentation, we opt for 13B models for both Vicuna v1.3 and Llama 2. In the LLM beam search sampling phase, we set the temperature parameter $t$ to $0.7$, and for each input sample, we accumulate the top $k=5$ LLM predictions for subsequent refinement. We activate the half-precision mode to enhance the efficiency of LLM generation. In order to preserve generation quality, quantization is not applied to the LLMs. The entirety of our experiments is performed with NVIDIA A100 GPUs, conducted in a single run. For TodSum dataset, the LLM generation time (from input to the end of beam search sampling) is 2.38/2.85 sec for Vicuna 1.3 and Llama 2 respectively. 

Regarding the BERT models integrated into the \mymethod pipeline, we select base-sized models for efficiency, namely RoBERTa-base (125M) and DeBERTa-v3-base-mnli (184M). The coefficients $\alpha, \beta, \gamma$ for final weighted scoring are tuned on separate validation sets, where a grid search is conducted with step size of 0.1. In uncertainty scoring, we found the size of nearest neighborhood $s = 3, 5$ generally lead to satisfactory performance in validation sets, and it is finally set to $5$ in all test settings. Note that after post-processing, the duplicate predictions are merged. A neighborhood of size $5$ may represent more than $5$ raw predictions. 

\section{Results and Analysis} 

\subsection{Effectiveness and Efficiency}

\begin{table*}[t!]

\Huge

\centering


\resizebox{\linewidth}{!}
{
\begin{tabular}{ll|cccc|cccc}
\hline
\textbf{}             &                                          & \multicolumn{4}{c|}{\textbf{TodSum}}                                           & \multicolumn{4}{c}{\textbf{DialogSum}}                                         \\
\textbf{Base LLM}     & \multicolumn{1}{c|}{\textbf{Refinement}} & \textbf{Rouge-1} & \textbf{Rouge-2} & \textbf{Rouge-L} & \textbf{BERTScore F1} & \textbf{Rouge-1} & \textbf{Rouge-2} & \textbf{Rouge-L} & \textbf{BERTScore F1} \\ \hline
\textbf{Vicuna v1.3}  & No                                       & 38.8             & 9.9              & 25.8             & 22.9                  & 32.7             & 10.3             & 26.3             & 27.6                  \\
\textbf{}             & Self-rerank                              & 39.1             & 10.2             & 25.6             & 22.9                  & 32.7             & 10.4             & 26.1             & 27.3                  \\
                      & \mymethod                               & \textbf{40.7}    & \textbf{11.1}    & \textbf{25.9}    & \textbf{24.5}         & \textbf{34.0}    & \textbf{10.9}    & \textbf{27.2}    & \textbf{28.5}         \\
                      & Oracle                                   & 45.7             & 13.0             & 29.6             & 28.6                  & 41.7             & 15.4             & 33.3             & 33.8                  \\ \hline
\textbf{Llama 2 chat} & No                                       & 40.1             & 10.3             & 26.4             & 23.2                  & 30.3             & 9.4              & 24.5             & 26.5                  \\
\textbf{}             & Self-rerank                              & 40.5             & 10.5             & 26.5             & 23.2                  & 30.4             & 9.7              & 24.6             & 26.3                  \\
                      & \mymethod                               & \textbf{41.4}    & \textbf{11.2}    & \textbf{26.7}    & \textbf{23.8}         & \textbf{30.8}    & \textbf{9.8}     & \textbf{25.0}    & \textbf{27.2}         \\
                      & Oracle                                   & 46.0             & 13.1             & 29.6             & 27.3                  & 38.0             & 13.2             & 30.6             & 31.2                  \\ \hline
\end{tabular}
}


\caption{Comparison of refinement methods for Abstractive Summarization tasks
}
\label{sum_main}
\end{table*}

\begin{table}[t!]

\Huge

\centering


\resizebox{\columnwidth}{!}
{
\begin{tabular}{llcc}
\hline
\textbf{Base LLM}     & \multicolumn{1}{c}{\textbf{Refinement}} & \textbf{TriviaQA} & \textbf{Natual Questions} \\ \hline
\textbf{Vicuna v1.3}  & No                                      & 57.8             & 18.6                      \\
\textbf{}             & Self-rerank                             & 60.0             & 19.4                      \\
\textbf{}             & Self-consistency                        & 59.7             & 20.0                      \\
                      & \mymethod                              & \textbf{62.0}    & \textbf{21.2}             \\
                      & Oracle                                  & 70.0             & 27.7                      \\ \hline
\textbf{Llama 2 chat} & No                                      & 51.0             & 15.3                      \\
\textbf{}             & Self-rerank                             & 51.4             & 15.4                      \\
\textbf{}             & Self-consistency                        & 51.6             & 15.7                      \\
                      & \mymethod                              & \textbf{55.1}    & \textbf{17.2}             \\
                      & Oracle                                  & 66.7             & 24.5                      \\ \hline
\end{tabular}
}


\caption{Comparison of refinement methods for Question Answering tasks
}
\label{qa_main}
\end{table}

\begin{figure}[ht]
\centering
\includegraphics[width=0.8\columnwidth,height=2in]{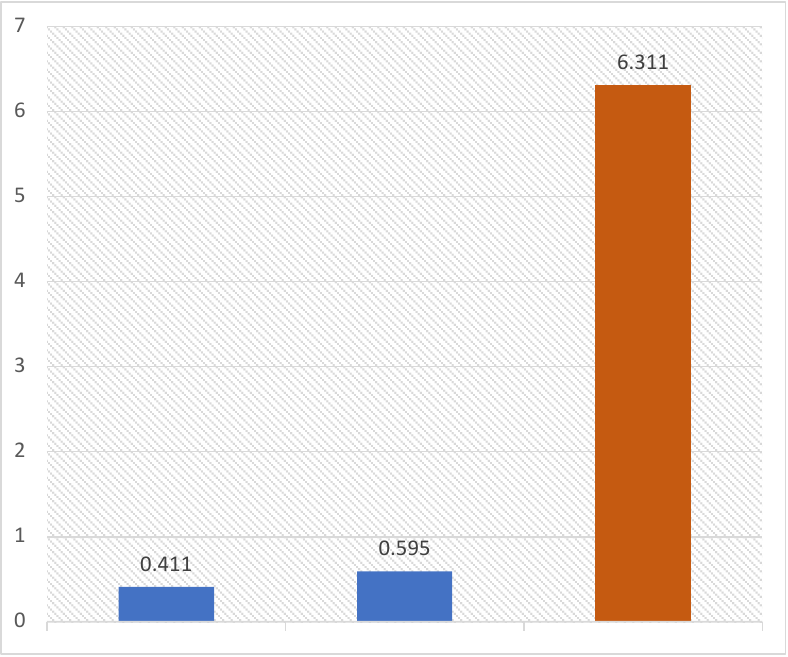}
\caption{Latency (sec) per input sample. From left to right: *BERT inference, \mymethod, and LLM self-rerank.}
\label{fig_latency}
\end{figure}

\textbf{Abstractive Summarization.} The experimental results for dialogue summarization are presented in Table~\ref{sum_main}. The initial performance of Vicuna v1.3 and Llama 2 chat in TodSum and DialogSum only result in moderate quality in generated summaries. However, the introduction of \mymethod brings obvious benefits into the enhancement of summarization outputs. Specifically, \mymethod achieves a decent improvement of 0.5-1.9 in Rouge-1 scores and 0.6-1.6 in BERTScore F1. 

The method consistently outperforms Self-rerank, emphasizing the significance of leveraging semantic stability, entailment, and inter-sample uncertainty measures in refining large language model generations.

\textbf{Question Answering.} As shown in Table~\ref{qa_main}, the baseline performance of the LLM on TriviaQA and Natural Questions reflects a gap between the difficulty of these two tasks. Despite the fact that they are both evaluated in closed-book setting, Natural Questions dataset has lower hit rate as it contains various challenging open-ended questions (e.g., Q: ``Philadelphia is known as the city of what?''. A: ``City of Brotherly Love'') Regardless of the challenges, \mymethod is able to improve upon no-refinement baseline for 1.6-4.2 points in hit rate, which consistently surpasses the both Self-rerank and Self-consistency approaches, indicating its effectiveness across diverse knowledge domains.

\begin{figure*}[ht]
\centering
\includegraphics[width=1.3\columnwidth,height=2in]{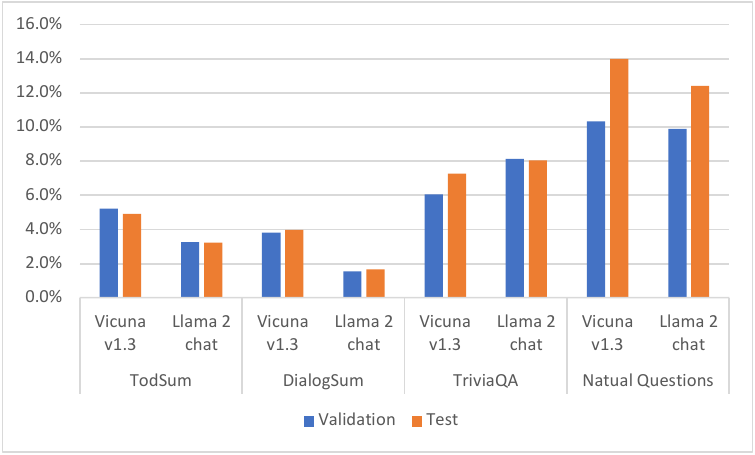}
\caption{Relative performance gains on validation and test sets. The best coefficient combination is tuned on validation sets. Evaluation metrics: Rouge-1 for TodSum and DialogSum, and hit rate for Trivia QA and Natural Questions.}
\label{fig_valid_test_relative}
\end{figure*}

\textbf{Inference Efficiency.}
Efficient inference is a crucial aspect of deploying language models in real-world applications. We analyze and compare the inference efficiency of the proposed \mymethod method against LLM Self-rerank. We use latency (in seconds) per input sample as a metric for assessing the efficiency of different inference pipelines. We report results on the TodSum dataset. Since the validation/test sets have size $ \leq 1000$, we use the entire sets instead of small batches for Inter-Sample Uncertainty scoring. 


As shown in Figure~\ref{fig_latency}, \mymethod exhibits remarkable efficiency advantages compared to LLM Self-rerank. The latency required by \mymethod is only 9.4\% of the latency observed in LLM Self-rerank\footnote{Both \mymethod and Self-rerank deal with predictions after LLM generation, and hence they don’t include the beam search sampling time.}. The majority of the latency in the \mymethod pipeline is attributed to *BERT inference, where *BERT refers to RoBERTa and DeBERTa models. The efficient integration of these models within the \mymethod framework contributes to its overall effectiveness while maintaining a significantly reduced latency compared to Self-rerank approaches. The efficiency improvement is particularly noteworthy, especially considering the demands of real-time applications where low latency is imperative.

\begin{table*}[t!]

\small

\centering


\begin{tabular}{ll|cccc}
\hline
\textbf{}            &                                          & \multicolumn{2}{c}{\textbf{Summarization - Rouge-1}} & \multicolumn{2}{c}{\textbf{QA - Hit rate}}   \\
\textbf{Base LLM}    & \multicolumn{1}{c|}{\textbf{Refinement}} & \textbf{TodSum}         & \textbf{DialogSum}         & \textbf{TriviaQA} & \textbf{Natual Questions} \\ \hline
\textbf{Vicuna v1.3} & No                                       & 38.78                   & 32.67                      & 57.82            & 18.61                     \\
\textbf{}            & Semantic stability only                  & 40.27                   & 34.17                      & 61.96            & 21.19                     \\
\textbf{}            & Entailment only                          & 40.67                   & 32.30                      & 57.66            & 18.51                     \\
\textbf{}            & Uncertainty only                         & 39.27                   & 32.60                      & 59.90            & 19.46                     \\
                     & \mymethod                               & 40.69                   & 34.27                      & 61.96            & 21.19                     \\ \hline
\end{tabular}


\caption{Ablation study of individual scoring dimensions
}
\label{ablation}
\end{table*}

\textbf{Overall Observations.} 
Figure~\ref{fig_valid_test_relative} provides a comprehensive overview of the relative performance improvement achieved on both validation and test sets. The test performance gains observed are generally on par with the validation settings and, in certain instances, even surpass them, as in the case of Natural Questions. This suggests that the weight tuning strategy employed during validation exhibits robustness and generalizability when applied to test sets. The potential explanation for larger gains in certain test cases could be attributed to the random split of test sets, providing certain sets with more room for improvement. 

The overall theme in the observed results is the consistently superior performance of \mymethod across all evaluated tasks. Furthermore, the consistency in performance gains between validation and test settings showcases the reliability and adaptability of the proposed \mymethod method across various settings in natural language processing tasks.

\subsection{Ablations and Hyperparameter Analysis}

\textbf{Ablations.} We systematically evaluate the impact of individual components within the \mymethod on both summarization and QA tasks, and present the findings in Table~\ref{ablation}. The results indicate that all three scoring dimensions have positive contributions in certain task scenario, compared to no-refinement baseline. Notably, semantic stability alone improves summarization Rouge-1 scores from 38.78 to 40.27 and 32.67 to 34.17 for TodSum and DialogSum respectively. Similarly, for question-answering, semantic stability increases the hit rate from 57.82 to 61.96, and 18.61.67 to 21.19, which are promising improvements.

Various NLP tasks have their own unique characteristics, suggesting that effectiveness of specific refinement dimensions might vary. For example, when considering the TodSum task, the nuances of entailment play a pivotal role in summarization quality for task oriented dialogues, where entailment scoring leads to the most significant gains. We further observed uncertainty scoring exhibits the best improvement in Appendix~\ref{appendix_additional_experimental_results}.

These insights underscore the synergies between semantic stability, entailment and uncertainty measures, highlighting their complementary roles in refining language model outputs. The comprehensive integration of these aspects in the \mymethod method showcases their collective impact, providing a flexible and contextually relevant refinement framework for various base LLMs and natural language processing tasks.

\begin{figure}[ht]
\centering
\includegraphics[width=0.9\columnwidth]{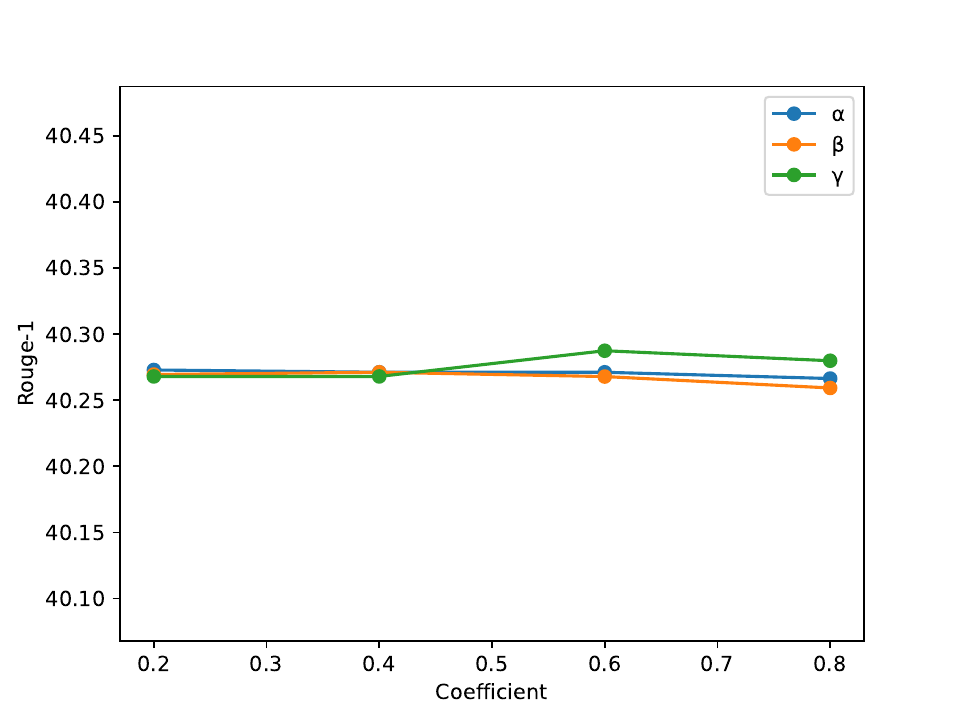}
\caption{Sensitivity analysis of coefficients for TodSum}
\label{fig_coeff}
\end{figure}

\textbf{Hyperparameter Analysis.} \label{hyperparameter_analysis} In Equation~\ref{eq_weighted}, coefficients $\alpha, \beta, \gamma$ are weights for three scoring dimensions respectively. To further investigate the sensitivity of non-trivial coefficients (when all coefficients are non-zero), a systematic approach was employed to assess the impact of individual coefficients on the overall model performance (Figure~\ref{fig_coeff}). A sensitivity analysis was conducted by keeping one coefficient, denoted as $x$, in a state of flux, while concurrently setting the other two coefficients, $y$ and $z$, to be 
$$y = z = \frac{1-x}{2}$$

Consider the green line in Figure~\ref{fig_coeff}: $\gamma$ is set as the variable, the relationship $\alpha = \beta = \frac{1 - \gamma}{2}$ is maintained. This variation allows for an in-depth exploration of the model's sensitivity to changes in each specific coefficient. illustrates that when all coefficients are non-zero, the model's performance remains relatively stable, with a fluctuation of Rouge-1 within 0.1, indicating the robustness to variations in individual coefficient values.

\section{Related Work}

\textbf{Prompting strategy for LLM improvement/refinement.} Improving LLM outputs by achieving behaviors close to reasoning has been explored before \cite{DBLP:conf/nips/Wei0SBIXCLZ22,DBLP:conf/iclr/0002WSLCNCZ23,yao2023tree,yao2023graph,DBLP:journals/corr/abs-2303-17651,yao2023react,gou2023critic,rl4f-2023}. \citet{DBLP:conf/nips/Wei0SBIXCLZ22} introduce a specific technique of formulating prompts for the model dubbed {\it Chain-of-Thought}. Essentially a series of intermediate reasoning steps that the model is asked to explicitly output, Chain of Thought significantly improves the ability of LLMs to perform complex reasoning.

\citet{DBLP:conf/iclr/0002WSLCNCZ23} propose a decoding strategy dubbed {\it Self-consistency}~--- under which the model, prompted in a chain-of-thought way, generates a set of sample predictions, or reasoning paths. The paths are then marginalized out, and the most consistent answer (the one which the most reasoning paths lead to) is selected as the final one. \citet{DBLP:journals/corr/abs-2210-11610} use this approach to improve LLMs without annotated data~--- they select the most consistent answer from the candidates pool, collect all the reasoning paths leading to that answer, and augment the trainset of the target model with the resulting data points.

In contrast to Self-consistency, the {\it Self-Refine} approach of \citet{DBLP:journals/corr/abs-2303-17651} assumes that the model iteratively provides verbal feedback on its own outputs, and incorporates it in the next generation round. Moreover, CRITIC~\cite{gou2023critic} empowers Language Models (LLMs) to independently verify and improve their own outputs with external toolkits, similar to the way humans make use of tools. All the approaches above require prompt engineering, while we tackle the problem from another perspective. 

\textbf{Model confidence/uncertainty without self-feedback.} A parallel line of work in improving LLM generation outputs is related to assessing the model confidence and the uncertainty of its predictions without iterative language model calls \cite{jiang-etal-2021-know,DBLP:conf/icml/LangAKS22,wang-etal-2022-uncertainty,cai2023kgeco,DBLP:conf/iclr/KuhnGF23,DBLP:conf/acl/GeLKG23,jiang-etal-2023-llm,vernikos2023small,he-etal-2024-zero,he2024semi,cai-etal-2023-masked}. \citet{DBLP:conf/iclr/KuhnGF23} define {\it semantic entropy}, a metric that incorporates linguistic invariance of the individual output candidates sharing identical meanings. This metric helps identify the correct model's predictions as evaluated on question answering task. 
LLM-Blender~\citet{jiang-etal-2023-llm} adopts a two-stage design, rank-and-fuse, to generate highly confident and superior candidate outputs. 
\citet{DBLP:conf/acl/GeLKG23} use uncertainty estimation in order to create modified pseudolabels, and define uncertainty of a pseudolabel (obtained using stochastic dropout-based model inference) as its proximity to other different pseudolabels for the same data point. 
Training on the selected pseudolabels increases performance in binary and multiclass classification, as well as Natural Language Understanding tasks. Selecting high-confidence pseudolabels is also a key aspect of the co-training technique proposed by \citet{DBLP:conf/icml/LangAKS22}, where both partial access and full access settings are studied. All these methods explore uncertainty/confidence from a certain perspective, while our approach combines the uncertainty/confidence with semantic stability and entailment, and we further proposed a framework for these three dimensions and investigated their synergies.

\section{Conclusions}
Our proposed \mymethod is an efficient framework to enhance text generation without the need for domain-specific training data or expensive annotations. By considering semantic stability, entailment, and inter-sample uncertainty measures, our approach significantly improves the quality of text generation across multiple natural language processing tasks. The efficiency and cost-effectiveness of our approach suggest its potential for wide adoption in real-world applications. 

\section{Limitations}
\mymethod can be potentially applied to a wide range of NLP problems, including dialogue response generation, open-ended common sense reasoning, and Natural Language Understanding (NLU) by text filling for text continuation. These topics require dedicated investigation and are not yet covered by this paper.

Our experiments show that beam search sampling almost always provides sufficient room for refinement, according to the oracle performance in Table~\ref{sum_main} and Table~\ref{qa_main}. Nevertheless, in certain task or data scenarios, performances of no-refinement baseline and oracle prediction may be close to each other. In that case, the performance of \mymethod will be limited by oracle results.

\newpage

\nocite{*} 
\bibliography{anthology,custom}
\bibliographystyle{acl_natbib}

\clearpage
\section*{Appendix}
\appendix

\section{Ethical Considerations}

We have reviewed all licenses of public datasets, which allow the usage for research and paper publication. All datasets are sets are de-identified to ensure anonymity.  

Our proposed method has a potential for substantial reductions in both the financial and environmental burdens associated with large language model improvement/refinement. Through minimizing the reliance on extensive data collection and human labeling, our approach serves as an effective safeguard for user and data privacy, mitigating the risk of information leakage during the construction of training corpora. 

During the paper writing process, Generative AI was only used for language checking, paraphrasing and polishing.

\label{appendix_additional_experimental_results}

\subsection{Claude v2}
\begin{table}[t!]

\Huge

\centering


\resizebox{\columnwidth}{!}
{

\begin{tabular}{lccc}
\hline
                        & \multicolumn{2}{c}{\textbf{DialogSum}}   & \textbf{TriviaQA}  \\
\textbf{Refinement}     & \textbf{Rouge-1} & \textbf{BERTscore F1} & \textbf{Hit rate} \\ \hline
No                      & 32.6             & 33.2                  & 74.2              \\
Semantic stability only & 34.8             & 35.2                  & 76.0              \\
Entailment only         & 35.0             & 35.3                  & 75.0              \\
Uncertainty only        & 35.0             & 35.7                  & \textbf{77.0}              \\
LLM-refine              & \textbf{36.9}    & \textbf{37.1}         & \textbf{77.0}     \\ \hline
\end{tabular}

}


\caption{Refinement performance on small test sets with Claude v2 as base LLM
}
\label{claude}
\end{table}

To further validate our approach using a closed-source model, we sampled small test sets (100 samples for each set), and conducted experiments with Anthropic Claude v2. 

As shown in Table~\ref{claude}, \mymethod achieves obvious improvements on both DialogSum and TriviaQA. In particular, we obtain improvements of 4.3 points in Rouge-1 on DialogSum and 2.8 points in hit rate on Trivia QA. These results indicate \mymethod is able to further improve the performance of closed-source/commercial-scale LLM. Interestingly, in contrast to Vicuna and Llama 2, uncertainty scoring yields the most competitive performance in ablations for both tasks, which shows different scoring dimentions in \mymethod has its own advantage in according to the model/task scenarios.



\section{Additional Sensitivity Analysis}

As described in \ref{hyperparameter_analysis}
{Hyperparameter Analysis}, the sensitivity analysis is conducted by keeping one coefficient, $x$ in a state of flux, while concurrently setting the other two coefficients, $y$ and $z$, to be $y = z = \frac{1-x}{2}$. Empirical results on four datasets show that the performance variations are very limited, with fluctuations of Rouge-1 within 0.2, and hit rate within 0.4. This indicates the stability of the method to variations in individual coefficient values.

\begin{figure*}[ht]
\centering
\includegraphics[width=16.5cm, height=4cm]{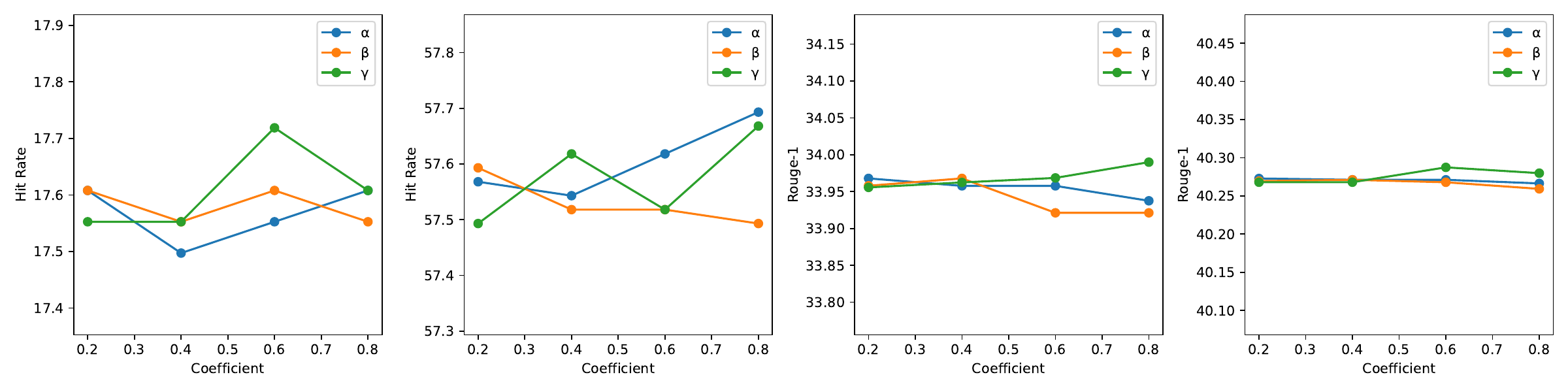}
\caption{Sensitivity analysis of coefficients for Natural Questions, TriviaQA, DialogSum and TodSum (from left to right in order)}
\label{fig_coeff_all_datasets}
\end{figure*}

\begin{table*}[t!]


\centering


\resizebox{\linewidth}{!}
{
\begin{tabular}{ll}
\hline
\multicolumn{1}{c}{\textbf{Input   dialogue}}                                                                                                                                                                                                                                                                                                                                                                                                                                                                                                                                                                                                                                                                                                                                                                                                                                                                                                                                                                                                                                             & \textbf{}      \\
\begin{tabular}[c]{@{}l@{}}Hi   there. Can you help me find a train from Norwich to Cambridge?\\      There are several trains on that route. What time would you like to   leave/depart?\\      I would like to leave after 12:15 on a Sunday.\\      I have a 12:16 departure from Norwich. Would you like me to book this for   you?\\      I'm not sure if I need to book just yet.    Can you tell me the travel time for this please?\\      Sure, TR5225 is a 79 minute trip and each seat costs 14.08 pounds.  Can I help you with anything else   today?\\      Is there an  All Saint's Church? I   would like to visit it as well.\\      Yes, all saints church is located t city cenre. It has free entrance.   \\      That sounds wonderful. Could I get the phone number along with the   attraction type?\\      It is listed as architecture, and their phone number is 012345678\\      Thank you so much! That is all the help I need for now.\\      You are very welcome! Have a great day and I hope you enjoy your visit!   Call for help anytime.\end{tabular} &                \\ \hline
\multicolumn{1}{c}{\textbf{LLM prediction}}                                                                                                                                                                                                                                                                                                                                                                                                                                                                                                                                                                                                                                                                                                                                                                                                                                                                                                                                                                                                                                               & \textbf{score} \\ \hline
\begin{tabular}[c]{@{}l@{}}A person is seeking help to book a train ticket from Norwich to Cambridge on   a Sunday at 12:16, \\      and the agent provides information on the travel time, ticket cost, and the   location of \teal{All Saints Church} in the city center, \\      with free entrance and a listed \teal{phone number} for architecture enthusiasts.\end{tabular}                                                                                                                                                                                                                                                                                                                                                                                                                                                                                                                                                                                                                                                                                                                      & 0.83           \\ \cline{1-1}
\begin{tabular}[c]{@{}l@{}}A person is seeking information about trains from Norwich to Cambridge and   asks about travel time, seat prices, \\      and the location of \teal{All Saints Church}, and the customer service   representative provides the necessary information and offers to book the   ticket.\end{tabular}                                                                                                                                                                                                                                                                                                                                                                                                                                                                                                                                                                                                                                                                                                                                                                    & 0.61           \\ \cline{1-1}
\begin{tabular}[c]{@{}l@{}}The text is a conversation between a person looking for train information   from Norwich to Cambridge and a train information provider, \\      who provides information on train schedules, fares, and nearby attractions.\end{tabular}                                                                                                                                                                                                                                                                                                                                                                                                                                                                                                                                                                                                                                                                                                                                                                                                                       & 0.12           \\ \hline
\multicolumn{1}{c}{\textbf{Ground   truth}}                                                                                                                                                                                                                                                                                                                                                                                                                                                                                                                                                                                                                                                                                                                                                                                                                                                                                                                                                                                                                                               &                \\
\begin{tabular}[c]{@{}l@{}}The user asks duration of the train. \\      This train leaves at 12:15 on sunday, departure is norwich and destination   is cambridge. \\      The user wants to know the \teal{phone number} and type of the attraction. the   attraction's name is \teal{All Saints Church}.\end{tabular}                                                                                                                                                                                                                                                                                                                                                                                                                                                                                                                                                                                                                                                                                                                                                                                 &                \\ \hline
\end{tabular}
}

\caption{Qualitative example: TodSum
}
\label{qualitative_todsum}
\end{table*}

\begin{table*}[t!]


\centering


\resizebox{\linewidth}{!}
{
\begin{tabular}{ll}
\hline
\multicolumn{1}{c}{\textbf{Input   question}}                                        & \textbf{}                                       \\
\multicolumn{2}{l}{In the novel, "Nicholas Nickelby", by Charles Dickens, what was the   name of the school, run by Wackford Squeers?} \\ \hline
\multicolumn{1}{c}{\textbf{LLM prediction}}                                          & \textbf{Score}                                  \\ \hline
\teal{Dotheboys Hall}                                                                       & 0.55                                            \\
Dotheboys                                                                            & 0.41                                            \\
Squeers School                                                                       & 0.29                                            \\ \hline
\multicolumn{1}{c}{\textbf{Ground   Truth}}                                          &                                                 \\
\teal{Dotheboys Hall}                                                                        &                                                 \\ \hline
\multicolumn{1}{c}{\textbf{Input question}}                                          &                                                 \\
\multicolumn{2}{l}{Who was the British Admiral who died in 1707 when four of his ships were   wrecked in the Scilly Isles?}            \\ \hline
\multicolumn{1}{c}{\textbf{LLM prediction}}                                          & \textbf{Score}                                  \\ \hline
\teal{Sir Cloudesley Shovell}                                                        & 0.68                                            \\
Russell                                                                              & 0.45                                            \\
Viscount Nicholas Boyle                                                              & 0.39                                            \\ \hline
\multicolumn{1}{c}{\textbf{Ground   Truth}}                                          &                                                 \\
{[}Cloudesley Shovell, \teal{Sir Cloudesley Shovell}{]}                                   &                                                 \\ \hline
\end{tabular}

}
\caption{Qualitative examples: TriviaQA
}
\label{qualitative_triviaqa}
\end{table*}


\section{Qualitative Analysis}

Selected qualitative examples from both TodSum and TriviaQA datasets are presented in Table~\ref{qualitative_todsum} and  Table~\ref{qualitative_triviaqa} respectively. Although the top $5$ beam search sampled candidates are considered in experiments, in Table~\ref{qualitative_todsum} and  Table~\ref{qualitative_triviaqa} we only present $3$ most representative predictions. 

In the example from TodSum, \mymethod is able to select the summary that contains key information ``All Saints Church'' and also mentions the phone number was provided. In fact, the prediction has the highest entailment score, which means it mostly implies other summaries. Regarding the examples from TriviaQA, the final score is largely determined by semantic stability: ``Dotheboys Hall'' and ``Dotheboys'' are close in semantic representation, and ``Sir Cloudesley Shovell'' (after parsing) actually appears $3$ times in top $5$ predictions.

\section{Prompt Templates}

\begin{table*}[t!]


\centering



\begin{tabular}{l}
\hline
\begin{tabular}[c]{@{}l@{}}

The following is a conversation between two individuals. Provide a brief summary \\ in [LENGTH] sentence(s). Output the summary only.   \\  

 \\  

Input example: [Input example]
 \\  
Output example: [Output example]  \\  

 \\  

Input: [INPUT] \\

\end{tabular} \\ \hline
\end{tabular}


\caption{Prompt template for Abstractive Summarization
}
\label{prompts_sum}
\end{table*}


\begin{table*}[t!]


\centering



\begin{tabular}{l}
\hline
\begin{tabular}[c]{@{}l@{}}

The following input is a question from an open domain Question-and-Answering task.  \\  Provide a succinct answer to the question in a single phrase (1-3 words). \\ In addition, provide supporting reasons step by step in the following format:  \\  

 \\  
 
Input example: Who was the first man to walk on the Moon? 

 \\  
Output example: Answer: Niel Armstrong. Reasoning: Neil Armstrong became the first human to walk  \\ on the moon  during NASA's Apollo 11 mission on July 20, 1969. This historic event is well-documented  \\ through photographs, videos, audio recordings, and historical records, providing irrefutable \\ evidence of his achievement.   \\  

 \\  

Input: [INPUT] 

\end{tabular} \\ \hline
\end{tabular}


\caption{Prompt template for QA
}
\label{prompts_qa}
\end{table*}


\begin{table*}[t!]


\centering



\begin{tabular}{l}
\hline
\begin{tabular}[c]{@{}l@{}}

The following input consists of generated predictions from a Large Language Model(LLM).  \\ Besides standard criteria like correctness and helpfulness, take semantic stability into account: \\ We prefer the candidate that is semantically closer to the majority of predictions. \\
Please choose exactly one best prediction, and output the item number (For example ``(8)''). \\
If there are multiple identical best answers, choose a random one. \\

\\

Input: [<TASK CONTEXT> Candidates: (1) ... (2) ... (k) ...]

\end{tabular} \\ \hline
\end{tabular}


\caption{Prompt template for LLM Self-rerank
}
\label{prompts_rerank}
\end{table*}

\label{appendix_prompts} The relevant prompt templates for Abstractive Summarization, QA and Self-rerank are presented in Table~\ref{prompts_sum}, Table~\ref{prompts_qa} and Table~\ref{prompts_rerank} respectively. We adopt Chain-of-Thought (CoT) prompting for open domain QA; while for Abstractive Summarization datasets, the key information is usually straightforward, hence we only include a length specification. Regarding LLM Self-rerank, we tested multiple additional instructions related to semantic stability, entailment/implication and uncertainty, and finally chose to include semantic stability only, as it produces most robust outcomes.

\end{document}